\newcommand{\CLASSINPUTtoptextmargin}{19mm}%
\newcommand{\CLASSINPUTbottomtextmargin}{43mm}%
\newcommand{\CLASSINPUTinnersidemargin}{12.9mm}%
\newcommand{\CLASSINPUToutersidemargin}{12.9mm}%
\documentclass[conference,10pt,a4paper]{IEEEtran}
\usepackage{amsmath}
\usepackage{times}
\usepackage{graphicx}
\usepackage{multirow}
\usepackage[none]{hyphenat}
\usepackage{float}
\usepackage{subfig}
\usepackage{iftex}
\usepackage{todonotes}
\usepackage{makecell}
\usepackage{mathtools}
\usepackage{gensymb}
\usepackage{caption} 
\restylefloat{table}

\newcommand{\CPcolumntwowidth}{21mm}%
\usepackage[shortcuts,acronym]{glossaries}
\newacronym{sc}{SC}{Silhouette score}
\newacronym{rbs}{RBS}{Rule-Based System}
\newacronym{fp}{FP}{False Positives}
\newacronym{fn}{FN}{False Negatives}
\newacronym{tp}{TP}{True Positives}
\newacronym{tn}{TN}{True Negatives}
\newacronym{ml}{ML}{Machine Learning}
\newacronym{nn}{NN}{Neural Network}
\newacronym{fmcw}{FMCW}{Frequency-Modulated Continuous Wave}
\newacronym{ai}{AI}{Artificial Intelligence}
\newacronym{id}{ID}{In-Distribution}
\newacronym{ood}{OOD}{Out-Of-Distribution}
\newacronym{ffnn}{FFNN}{Feed-Forward Neural Network}
\newacronym{iai}{IAI}{Interpretable AI}
\newacronym{shap}{SHAP}{Shapley Additive Explanations}

\newacronym[\glsshortpluralkey={NNs}]{NN}{NN}{Neural Network}

\usepackage{amssymb}
\usepackage[backend=bibtex,defernumbers=true,sorting=none,style=ieee]{biblatex}%
\ifPDFTeX%
\usepackage{t1enc}
\usepackage{times}
\fi%
\ifXeTeX%
\usepackage{fontspec}
\fi%
\ifLuaTeX%
\usepackage{fontspec}
\fi%
\makeatletter

\def\@maketitle{\newpage
\bgroup\par\addvspace{0.5\baselineskip}\centering%
\ifCLASSOPTIONtechnote
   {\bfseries\large\@IEEEcompsoconly{\sffamily}\@title\par}\vskip 1.3em{\lineskip .5em\@IEEEcompsoconly{\sffamily}\@author
   \@IEEEspecialpapernotice\par{\@IEEEcompsoconly{\vskip 1.5em\relax
   \@IEEEtitleabstractindextextbox{\@IEEEtitleabstractindextext}\par
   \hfill\@IEEEcompsocdiamondline\hfill\hbox{}\par}}}\relax
\else
   \vskip0.2em{\EuMWtitlesize\ifCLASSOPTIONtransmag\bfseries\LARGE\fi\@IEEEcompsoconly{\sffamily}\@IEEEcompsocconfonly{\normalfont\normalsize\vskip 2\@IEEEnormalsizeunitybaselineskip
   \bfseries\Large}\@title\par}\vskip1.0em\par
   \ifCLASSOPTIONconference%
      {\@IEEEspecialpapernotice\mbox{}\vskip\@IEEEauthorblockconfadjspace%
       \mbox{}\hfill\begin{@IEEEauthorhalign}\@author\end{@IEEEauthorhalign}\hfill\mbox{}\par}\relax
   \else
      \ifCLASSOPTIONpeerreviewca
         {\@IEEEcompsoconly{\sffamily}\@IEEEspecialpapernotice\mbox{}\vskip\@IEEEauthorblockconfadjspace%
          \mbox{}\hfill\begin{@IEEEauthorhalign}\@author\end{@IEEEauthorhalign}\hfill\mbox{}\par
          {\@IEEEcompsoconly{\vskip 1.5em\relax
           \@IEEEtitleabstractindextextbox{\@IEEEtitleabstractindextext}\par\hfill
           \@IEEEcompsocdiamondline\hfill\hbox{}\par}}}\relax
      \else
         \ifCLASSOPTIONtransmag
           {\@IEEEspecialpapernotice\mbox{}\vskip\@IEEEauthorblockconfadjspace%
            \mbox{}\hfill\begin{@IEEEauthorhalign}\@author\end{@IEEEauthorhalign}\hfill\mbox{}\par
           {\vspace{0.5\baselineskip}\relax\@IEEEtitleabstractindextextbox{\@IEEEtitleabstractindextext}\vspace{-1\baselineskip}\par}}\relax
         \else
           {\lineskip.5em\@IEEEcompsoconly{\sffamily}\sublargesize\@author\@IEEEspecialpapernotice\par
           {\@IEEEcompsoconly{\vskip 1.5em\relax
            \@IEEEtitleabstractindextextbox{\@IEEEtitleabstractindextext}\par\hfill
            \@IEEEcompsocdiamondline\hfill\hbox{}\par}}}\relax
         \fi
      \fi
   \fi
\fi\par\addvspace{0.0\baselineskip}\egroup}

\def\EuMWtitlesize{\@setfontsize{\EuMWtitlesize}{24}{24pt}}
\def\EuMWauthorsize{\@setfontsize{\EuMWauthorsize}{11}{11pt}}
\def\EuMWaffilsize{\@setfontsize{\EuMWaffilsize}{10}{10pt}}
\def\EuMWcaptionsize{\@setfontsize{\EuMWcaptionsize}{9}{10pt}}
\def\EuMWbibsize{\@setfontsize{\EuMWbibsize}{8}{10pt}}

\def\@IEEEauthorblockNstyle{\EuMWauthorsize\@IEEEcompsocnotconfonly{\sffamily}\@IEEEcompsocconfonly{\large}}
\def\@IEEEauthorblockAstyle{\EuMWaffilsize\@IEEEcompsocnotconfonly{\sffamily}\@IEEEcompsocconfonly{\itshape}\@IEEEcompsocconfonly{\large}}
\def\@IEEEauthordefaulttextstyle{\EuMWauthorsize\@IEEEcompsocnotconfonly{\sffamily}\sublargesize}

\def\thebibliography#1{\section*{\refname}%
    \addcontentsline{toc}{section}{\refname}%
    \EuMWbibsize\@IEEEcompsocconfonly{\small}\vskip 0.3\baselineskip plus 0.1\baselineskip minus 0.1\baselineskip
    \list{\@biblabel{\@arabic\c@enumiv}}%
    {\settowidth\labelwidth{\@biblabel{#1}}%
    \leftmargin\labelwidth
    \advance\leftmargin\labelsep\relax
    \itemsep \IEEEbibitemsep\relax
    \usecounter{enumiv}%
    \let\p@enumiv\@empty
    \renewcommand\theenumiv{\@arabic\c@enumiv}}%
    \let\@IEEElatexbibitem\bibitem%
    \def\bibitem{\@IEEEbibitemprefix\@IEEElatexbibitem}%
\def\newblock{\hskip .11em plus .33em minus .07em}%
\ifCLASSOPTIONtechnote\sloppy\clubpenalty4000\widowpenalty4000\interlinepenalty100%
\else\sloppy\clubpenalty4000\widowpenalty4000\interlinepenalty500\fi%
    \sfcode`\.=1000\relax}

\long\def\@makecaption#1#2{%
\ifx\@captype\@IEEEtablestring%
\par\@IEEEtabletopskipstrut
\else
\@IEEEfigurecaptionsepspace
\fi
\setbox\@tempboxa\hbox{\normalfont\footnotesize {#1.}\nobreakspace\nobreakspace #2}%
\ifdim \wd\@tempboxa >\hsize%
\setbox\@tempboxa\hbox{\normalfont\footnotesize {#1.}\nobreakspace\nobreakspace}%
\parbox[t]{\hsize}{\normalfont\footnotesize\noindent\unhbox\@tempboxa#2}%
\else
\ifCLASSOPTIONconference \hbox to\hsize{\normalfont\footnotesize\hfil\box\@tempboxa\hfil}%
\else \hbox to\hsize{\normalfont\footnotesize\box\@tempboxa\hfil}%
\fi\fi
\ifx\@captype\@IEEEtablestring%
\@IEEEtablecaptionsepspace
\else
\fi}

\newlength\tablecaptiontotableskip
\newlength\figuretocaptionskip
\def\@IEEEfigurecaptionsepspace{\vskip\figuretocaptionskip\relax}%
\def\@IEEEtablecaptionsepspace{\vskip\tablecaptiontotableskip\relax}%

\def\abstract{\normalfont%
\@IEEEabskeysecsize\bfseries\textit{\abstractname}\,\bfseries\textit{---}\,%
\@IEEEgobbleleadPARNLSP}%

\def\IEEEkeywords{\normalfont%
\@IEEEabskeysecsize\bfseries\textit{\IEEEkeywordsname}\,\bfseries\textit{---}\,%
\@IEEEgobbleleadPARNLSP}%
\def\endIEEEkeywords{\relax\vspace{0.67ex}%
\par\if@twocolumn\else\endquotation\fi%
\normalsize\normalfont}%

\DeclareRobustCommand*{\EuMWauthorrefmark}[1]{\raisebox{0pt}[0pt][0pt]{\textsuperscript{#1}}}%
\def\@IEEEauthorblockNtopspace{0ex}
\def\@IEEEauthorblockAtopspace{1mm}
\def\tablename{Table}
\def\thetable{\arabic{table}}
\def\IEEEkeywordsname{Keywords}
\def\subsubsection{\@startsection{subsubsection}{3}{\z@}{1.5ex plus 1.5ex minus 0.5ex}%
{0.7ex plus .5ex minus 0ex}{\normalfont\normalsize\itshape}}%
\newlength{\CPheadmatchindent}%
\def\@seccntformat#1{\hbox to\CPheadmatchindent{\csname the#1dis\endcsname}\hskip 0.1em \relax}
\IEEEilabelindentA \parindent
\IEEEilabelindent \IEEEilabelindentA
\IEEEelabelindent \parindent
\IEEEdlabelindent \parindent
\IEEElabelindent \parindent
\makeatother

\begin{document}
\raggedbottom
%
%
%
\title{Interpretable Rule-Based System for Radar-Based Gesture Sensing: Enhancing Transparency and Personalization in AI}
%
%
\author{%
\IEEEauthorblockN{%
Sarah Seifi\EuMWauthorrefmark{\#\$}, 
Tobias Sukianto\EuMWauthorrefmark{\$$*$}, 
Cecilia Carbonelli\EuMWauthorrefmark{\$}, 
Lorenzo Servadei\EuMWauthorrefmark{\#},
Robert Wille\EuMWauthorrefmark{\#$\wedge$}
}
\IEEEauthorblockA{%
\EuMWauthorrefmark{\#}Technical University Munich, Germany\\
\EuMWauthorrefmark{\$}Infineon Technologies AG, Germany\\
\EuMWauthorrefmark{$*$}Johannes Kepler University Linz, Austria\\
\EuMWauthorrefmark{$\wedge$}Software Competence Center Hagenberg GmbH (SCCH), Austria\\
\EuMWauthorrefmark{}sarah.seifi@tum.de\\}}

\maketitle
%
%
%
\begin{abstract}
The increasing demand in artificial intelligence (AI) for models that are both effective and explainable is critical in domains where safety and trust are paramount. In this study, we introduce MIRA, a transparent and interpretable multi-class rule-based algorithm tailored for radar-based gesture detection. Addressing the critical need for understandable AI, MIRA enhances user trust by providing insight into its decision-making process. We showcase the system's adaptability through personalized rule sets that calibrate to individual user behavior, offering a user-centric AI experience. Alongside presenting a novel multi-class classification architecture, we share an extensive frequency-modulated continuous wave radar gesture dataset and evidence of the superior interpretability of our system through comparative analyses. Our research underscores MIRA's ability to deliver both high interpretability and performance and emphasizes the potential for broader adoption of interpretable AI in safety-critical applications.
\end{abstract}
%
%

\section{Introduction}
\label{sec:introduction}

In a time where the importance of transparency within \ac{ai} is ever-increasing, our research presents an innovative \ac{rbs} explicitly designed for radar-based gesture sensing. This contrasts sharply with the prevalent black-box \glspl{nn}, which, despite their effectiveness, often fail to provide clarity in decision-making processes. In sectors where interpretability is critical, such as healthcare and security, this raises concerns \cite{wang2022medical}.
Radar-based gesture-sensing offers unique advantages to common camera-based sensors, particularly in scenarios where visual or tactile interaction is limited. This technology enables hands-free interactions, making it ideal for automotive, healthcare, and smart home applications. However, existing gesture-sensing approaches often rely on complex NNs \cite{kabisha2022face}, hindering interpretability and customization. 
To meet the growing demand for personalized experiences, there is a pressing need for \ac{ml} models that can adapt to individual users' behavior and characteristics.
In response, we propose an interpretable and personalizable \underline{\textbf{m}}ult\underline{\textbf{i}}-class \underline{\textbf{r}}ule-based \underline{\textbf{a}}lgorithm (\textbf{MIRA}), which employs rule induction to automatically generate transparent if-then rules. Each rule, potentially composed of multiple conditions known as literals, uncovers meaningful patterns within the data and shows the reasoning behind the system's decisions.

MIRA is specifically tailored for sensing applications and is applied to \ac{fmcw} radar-based gesture sensing.
Distinguishing our algorithm from conventional \ac{rbs}s like RIPPER \cite{ripper}, IREP \cite{irep}, SkopeRules \cite{skoperules}, and FOLD-R++ \cite{wang2022fold}, MIRA can perform multi-class (gesture) classification. Past models have predominantly focused on binary classification tasks, neglecting the complexities inherent
in multi-class scenarios commonly encountered in gesture classification applications.
Additionally, they obscure interpretability through the use of non-interpretable \ac{ml} models like Random Forests or heuristics, e.g., information gain and the Gini impurity index for rule development. Our approach, on the other hand, harnesses a comprehensible combination of a weighted \ac{sc} and the F-Beta score. This approach not only promotes interpretability but also refines the rules for each gesture class, ensuring that significant features are not overlooked and providing a more balanced assessment across different classes.
Additionally, we innovate through the introduction of personalized rule systems. These personalized rules enhance the model's adaptability, allowing it to adjust to the nuances of individual user behaviors and preferences, thereby tailoring the user experience.

We present a novel methodology adapted for \ac{fmcw} radar-based gesture sensing. This approach harnesses the strengths of \ac{rbs}s to capture relationships in radar data, translating them into simple yet informative rules for distinguishing between five gesture classes.

Our contributions include: 1. Introducing MIRA, a new \ac{rbs} architecture based on the F-Beta score and a novel weighted \ac{sc} for multi-class (radar-based gesture) classification, expanding beyond previous algorithms only supporting binary classification. 2. Development of personalized rule systems based on foundational rules, enhancing adaptability and performance. 3. Publication of an extensive \ac{fmcw} radar-based gesture-sensing dataset containing 21k gestures \cite{dataset}.

\section{Rule-Based Gesture Sensing}
\label{sec:page style}


This section outlines the architecture of MIRA, our \ac{rbs} for gesture sensing, covering the rule induction process and the integration of personalized rules for improved user adaptability.



\subsection{Algorithm for Rule-Based Gesture Sensing}

\textbf{Rule Induction Based on Sequential Covering}. In this work, we employ a rule induction process based on sequential covering to develop the \ac{rbs} for interpreting radar-based gesture sensing. 
MIRA iteratively selects the best rule to add to an initially empty rule set and hence sequentially removes samples that the rule covers until (almost) all samples have been accounted for or a stopping criterion is met. Our algorithm develops and optimizes rules based on a training and a validation dataset, respectively, using rule-stopping and early-stopping criteria. 
The primary criteria involve setting a maximum number of rules and literals, to maintain the complexity of the rule set. Other criteria are based on the coverage and the accuracy of a rule. The former refers to the fraction of samples that satisfy the rule's conditions, whereas the latter is the fraction of correctly covered samples.
In this work, we set the minimum coverage of the newly developed rule on the training dataset. Additionally, minimum rule coverage and accuracy rate are defined on a validation dataset. As a last stopping criterion, a minimum number of samples in the training and the validation dataset are defined. If the number of samples left in each dataset is less than the specified values, the rule development is stopped. These measures are put in place to prevent overfitting rules on a small number of samples and outliers in the training dataset. On top of that, they ensure generalizability by evaluating the rule on a separate validation dataset.
These stopping criteria balance model complexity and performance, ultimately reducing the number of conditions and overall rules.

\textbf{Weighted Silhouette and F-Beta Score for Rule Development.} 
To enable multi-class classification, we identify the most compact gesture class for rule development based on a weighted \ac{sc}. This is the first step of MIRA. The \ac{sc} of a cluster is computed based on the average distances within and outside the cluster, providing an intuitive alternative to traditional methods like information gain. This score ranges from $-1$ to $1$, where a score closer to $1$ indicates that the object is well-matched to its own and poorly matched to neighboring clusters. The \ac{sc} of a gesture class $j$ is defined as:
\begin{equation}
    SC_j = \frac{1}{N}\sum_{i=1}^{N}\frac{\text{b}(\textbf{x}_i)-\text{a}(\textbf{x}_i)}{\text{max}\{\text{a}(\textbf{x}_i), \text{b}(\textbf{x}_i)\}}
\end{equation}
with $N$ being the total number of samples. $\text{a}(\textbf{x}_i)$ is the average distance of the gesture sample $\textbf{x}_i \in \mathbb{N}^{n_{\text{frames}} \times n_{\text{features}}}$, with $n_{\text{frames}}$ being the number of frames and with $n_{\text{features}}$ being the number of features, to all other samples in the same cluster. $\text{b}(\textbf{x}_i)$ is the average distance of $\textbf{x}_i$ to the samples in other clusters.

To prevent biased rule development towards classes with fewer samples, a weighted \ac{sc} method, incorporating cluster size and sample distribution, is proposed. This approach aims to avoid favoring smaller clusters with a higher \ac{sc}, ensuring a fair evaluation of all classes.
The weighted \ac{sc} developed in this work is:
\begin{equation}
    SC_{\text{weighted}, j} = \lambda_1 \sqrt{\lambda_2 \frac{n_j}{N_{\text{left}}}} + \lambda_3  \cdot SC_j
\end{equation}
with $\lambda_1 \in [0,1]$ weighting the overall transformed impact based on the cluster size and $\lambda_2 \in [0,\infty)$ directly weighting $n_j$, the number of samples in the current cluster $j$ relative to all not yet classified samples $N_{\text{left}}$ remaining in the training dataset. $\lambda_3 \in [0,1]$ adjusts the weight of $SC_j$.

Once the optimal gesture class is identified, the second step of MIRA is started, which is rule development and evaluation using the F-Beta score. It combines precision and recall into a single metric, with the flexibility of adjusting the importance ratio $\beta \in [0,\infty)$: 
\begin{equation}
    \begin{split}
         \text{F-Beta} &= (\beta^2 + 1)\frac{\text{Precision}\cdot \text{Recall}}{\beta^2\text{Precision}+\text{Recall}} \\
        &= (\beta^2+1)\frac{TP}{\beta^2(FN+FP)+(1-\beta^2)TP}.
    \end{split}
\end{equation}

The F-Beta score is useful when both precision and recall are taken into account, but their respective weights should be adjusted. In this work, we prioritize \ac{fp}, i.e., precision, over \ac{fn}, i.e., recall by setting $\beta < 1$ to $0.3$. Falsely negative predicted gestures, which are not being covered by a rule, can be correctly classified by a subsequent rule, whereas the opposite is not achievable. The algorithm allows for dynamic evaluation and refinement of rule literals based on the F-Beta score, providing control over the rule-generation process. 

\begin{figure}[b!]
\centering
       \includegraphics[width = 1\linewidth]{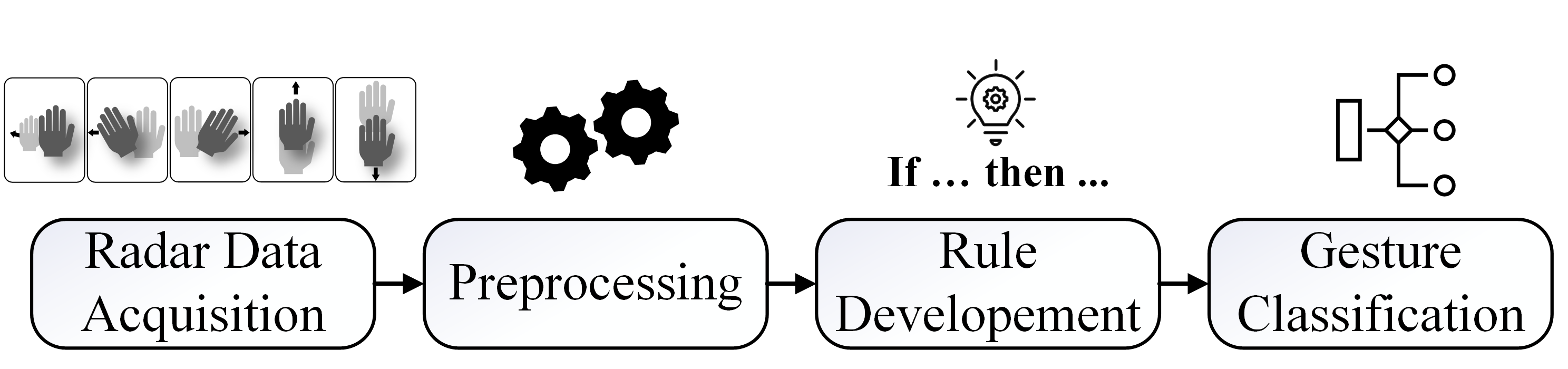}
      \vspace{-3mm}
    \caption{Algorithm's progression from data capture to gesture recognition.}
    \vspace{-3mm}
\label{fig:flow}
 \end{figure}

As a last step, upon meeting one of the defined stopping criteria, a default-else rule based on majority voting is incorporated into the rule set. This final rule is assigned to the most prevalent gesture class among the samples not accounted for by the preceding rules. Fig.\,\ref{fig:flow} illustrates the main components of the rule-based gesture sensing algorithm.

\subsection{Foundational and Personalized Rules}

In the realm of gesture sensing, the performance gap between a \ac{rbs} and more advanced techniques like black-box NNs is evident. The latter can achieve superior results. The former is constrained by its inability to capture the complex dynamics of gestures across diverse individuals and environments, limiting its generalization abilities even with validation sets and early-stopping criteria.

To overcome these constraints, we propose an innovative approach centered on foundational and personalized rule sets. Foundational rules provide a broad understanding of gesture sensing applicable to all users. They are derived from diverse training data encompassing various individuals with distinct gesture execution patterns and environmental conditions. Personalized rules, tailored to individual users, accommodate unique gesture execution nuances stemming from anatomical differences, motor skills, personal preferences, and even cultural influences.

Using user-specific data, personalized rules refine foundational rules, enhancing adaptability and accuracy. Users perform a set number of gestures for each class, with those not captured by the last default-else rule used to iterate over the \ac{rbs} algorithm, generating personalized rules. This iterative process enables the identification of user-specific rules tailored to individual gesture executions, which are subsequently appended to the foundational rules.

Experiments with varying calibration gestures validate the efficacy of personalized rules, detailed in Section\,\ref{subsec:user-centric-rbs}.

\section{Experiments and Discussion}
This section presents MIRA's performance in multi-class gesture classification with our radar dataset, contrasts it with conventional \ac{rbs}s and a \ac{ffnn}, and highlights the interpretability and customization through foundational and personalized rules.

\subsection{Radar System and Dataset Description}
In this research, Infineon Technologies' XENSIV™ BGT60LTR13C $60\,$GHz \ac{fmcw} radar has been used. This radar was set to function across a frequency range from $58.5\,$GHz to $62.5\,$GHz, providing a range resolution of $37.5\,$mm and a maximum resolvable range of $1.2\,$m. For the transmission of signals, it uses a burst configuration, emitting $32$ chirps within each burst at a frame rate of $33.3\,$Hz, and a pulse repetition time of $300\,\mu$s.

A total of $21,000$ gestures were performed by eight individuals in six different locations, with a field view of $\pm \text{ }45 \degree$ and a distance of $\le 1\,$m from the radar. The dataset includes five different gesture types, including \textit{Swipe Left}, \textit{Swipe Right}, \textit{Swipe Up}, \textit{Swipe Down}, and \textit{Push}. A recording has an approximate duration of $3\,$s or 100 frames with the gesture execution itself being $0.3\,$s or ten frames long. The dataset is accessible on IEEE Dataport \cite{dataset}.

Since this work is centered on the interpretability of radar-based \ac{ai}, we utilize the lightweight radar processing algorithm introduced in \cite{Strobel_2024}. The hand movements are characterized via five features, i.e., \textit{radial distance} (range), \textit{radial velocity} (Doppler), \textit{horizontal angle} (azimuth), \textit{vertical angle} (elevation), and \textit{signal magnitude} (peak). For the input to the \ac{ml} models, we distilled these features by averaging their values across only the ten gesture frames.

\subsection{Comparative Evaluation of Rule-Based Systems}
This section details a comparative analysis that reveals MIRA's enhanced performance, particularly in multi-class classification tasks, as outlined in Table\,\ref{tab:RBS-Benchmarking}. This capability is a notable advancement over conventional \ac{rbs} frameworks, which typically do not support multi-class functionality.
The evaluation protocol was rigorous, involving experiments not only on the training data but also on both \ac{id} and \ac{ood} datasets, incorporating data from a variety of users and environments. \ac{id} and \ac{ood} data refer to matching and non-matching distributions with the training data, respectively.
The dataset of $21,000$ gestures were split into $11,000$ gestures based on four users in three locations and into $10,000$ remaining gestures. From the $11,000$ gestures, $7,920$ and $1,980$ gestures were used as training and validation data, respectively. $1,100$ recordings were used as \ac{id} test data whereas the remaining $10,000$ were used as \ac{ood} test data.

To facilitate a fair comparison with binary classification \ac{rbs}s, we developed individual rule sets for each class. These were then merged to construct a comprehensive rule set designed for multi-class classification. To ensure comparability, we standardized the number of rules and literals across systems to a limit of 15 rules with a maximum of two literals each. For MIRA, the minimum rule coverage on the training dataset was set to eight, and on the validation dataset was set to five samples. The minimum validation rule accuracy rate for rule evaluation was set to $70\%$. If less than six or two samples were left in the training and validation dataset, respectively, early stopping was triggered to avoid overfitting. The empirically found parameters of the weighted \ac{sc} were always set to $\lambda_1=0.5$, $\lambda_2=10$, and $\lambda_3=0.7$.

Our findings show that MIRA not only attains higher accuracy in real-world multi-class scenarios (accuracy increase up to $13.4 \%$ for \ac{id} $14\%$ for \ac{ood} data) but also provides a more streamlined and intuitive set of rules. Instead of needing different rule sets for each gesture class, we can provide an interpretable rule list capable of multi-class classification. This leads to greater model transparency, meeting the growing need for \ac{ai} systems that are both effective and easy to understand.


{

\begin{table}[t]
\caption{Comparison of RBS implementations: Performance evaluation.}
\small
\centering
\begin{tabular}{|l|l|l|l|}\hline

\makecell{Algorithms} & \makecell{Train Acc.{[}\%{]}}  & \makecell{ID Acc.{[}\%{]}} & \makecell{OOD Acc.{[}\%{]}} \\ \hline
SkopeRules & \parbox[t]{\CPcolumntwowidth}{\strut76.6 }& 77.4& 75.5 \\ \hline
RIPPER & 76.1 & 75.8 & 69.9 \\ \hline
IREP & 81.1 & 80.4 & 79.8 \\ \hline
MIRA & \textbf{88.4}& \textbf{89.2}& \textbf{83.9}\\ \hline
\end{tabular}
\label{tab:RBS-Benchmarking}
\end{table}
}

\subsection{Performance-Interpretability Trade-off in Radar-based Gesture Sensing}



This section emphasizes the trade-off between performance and interpretability when choosing between an \ac{rbs} and a \ac{ffnn} for radar-based gesture sensing. 
The \ac{ffnn} consists of three fully connected layers. The first two layers each have 16 nodes and a ReLU activation function, while the third layer has five nodes (corresponding to the number of classes) and a softmax activation function. During training, the model uses the Adam optimizer, a learning rate of $0.001$, the sparse categorical cross-entropy loss function, and a batch size of 32 for 100 epochs. 

\begin{figure}[tb!]
\centering
       \includegraphics[width = 1\linewidth]{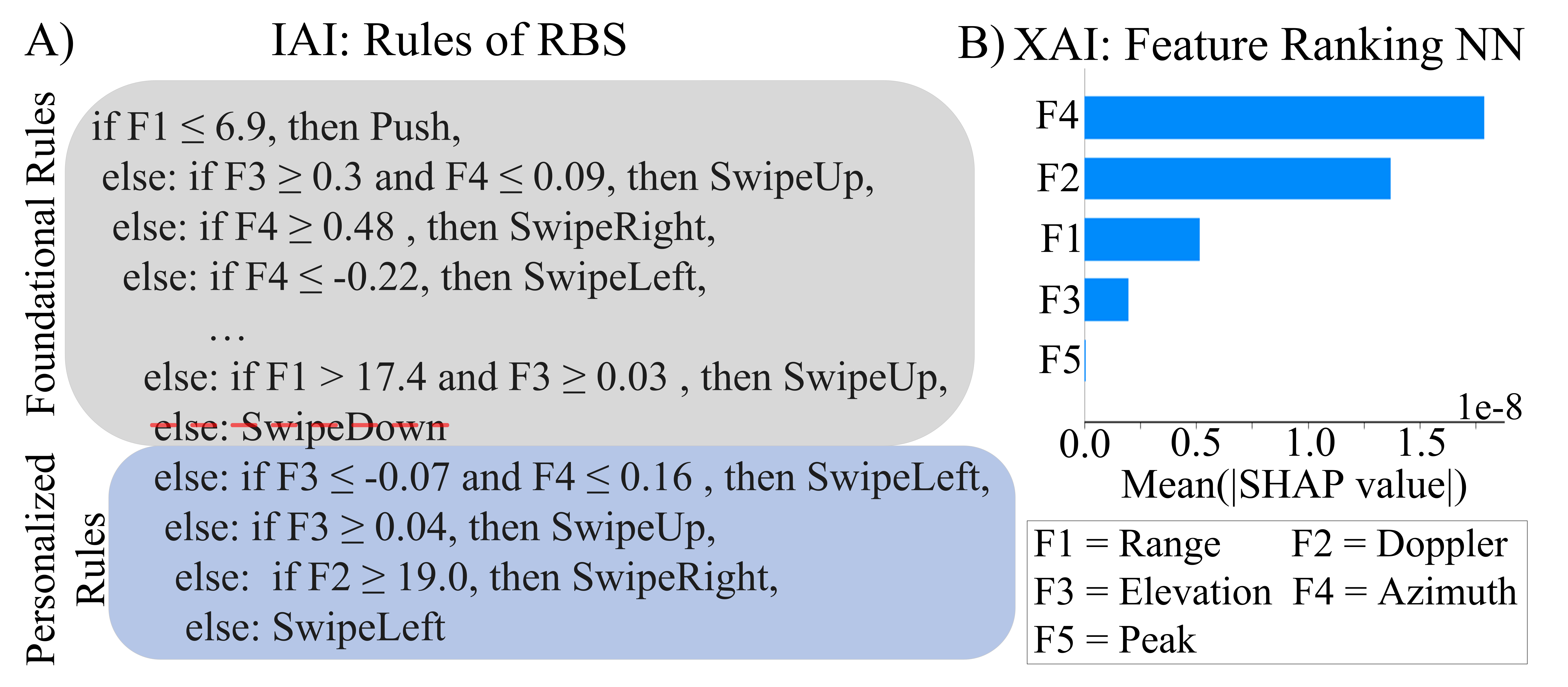}
      \vspace{-3mm}
    \caption{Comparing model interpretability: A) MIRA generates transparent rules. After removing the last else-default rule, personalized rules are appended. B) XAI methods provide insights into the model, highlighting feature influences. Full transparency is not achieved.}
    \vspace{-3mm}
\label{fig:xai-vs-iai}
 \end{figure}

Although the FFNN outperforms the foundational RBS in terms of \ac{ood} accuracy ($91\%$ vs. $83.9\%$), its decision-making process lacks transparency. In contrast, MIRA excels in providing clear insights into its decision-making (s. Fig.\,\ref{fig:xai-vs-iai}\,A)), making it preferable when model understanding is paramount.
To bridge the gap between performance and transparency, we applied an explainable AI technique called \ac{shap} to the \ac{ffnn} \cite{lundberg2017unified}. \ac{shap} helps to explain the \ac{ffnn}'s decision-making by ranking the importance of input features based on their contribution to the model's predictions. Despite the insights provided by \ac{shap}, comprehending the decision-making process of the FFNN remains more intricate compared to interpreting the rule set of MIRA. As depicted in Fig.\,\ref{fig:xai-vs-iai}\,B), the feature importances for \textit{SwipeLeft} and \textit{SwipeRight} gestures are shown, highlighting the pronounced impact of the azimuth feature on the model output, aligning with the execution of the gestures. While \ac{shap} provides a hierarchy of feature relevances, the clarity and direct interpretability of MIRA's rules are still unmatched. They allow for complete transparency, ensuring that the rationale behind each predicted gesture class is fully comprehensible.

{


\subsection{Foundational and Personalized Rules for Improved Radar-Based Gesture Sensing}
\label{subsec:user-centric-rbs}

By introducing the concept of foundational and personalized rules, as highlighted in Fig.\,\ref{fig:xai-vs-iai}\,A), the study achieved higher model performance while still maintaining interpretability and allowing model customization. As shown in Table\,\ref{tab:found-rules}, the model accuracy for six different users with and without personalized rules was assessed. The removal of the last else-default rule and the addition of up to four personalized rules to the foundational rules improved model performance (average performance boost of up to $9.6 \%$). The user pattern may not be captured sufficiently with fewer rules and may lead to overfitting with more rules. The study found that a small set of personalized rules, developed using only five gestures per gesture class, was sufficient to achieve this performance improvement.

\begin{table}[t]
\small
\centering
\begin{tabular}{|l|l|l|l|l|l|l|}\hline
\caption{MIRA's accuracy of foundational and personalized rules using different numbers of gestures for calibration.}
\multirow{2}{5mm}{\parbox{8mm}{{\bfseries User}}} &\multirow{2}{8mm}{\parbox{7mm}{{\bfseries Found. rules [\%]}}}& \multicolumn{5}{c|}{\raisebox{-2mm}{\bfseries Person. rules with \textit{n} calibration gestures [\%]}}\\ \cline{3-7}
 &  & \makecell{\textit{n} = 200} & \makecell{\textit{n} = 100} & \makecell{\textit{n} = 50} & \makecell{\textit{n} = 10} & \makecell{\textit{n} = 5} \\ \hline
1 & 84.5& 92.5& 89.1 & 88.4 & 88.2 & 87.8 \\ \hline
2 & 80.6& 92.1& 90.2 & 90 & 89.1 & 88.3 \\ \hline
3 & 80.0 & 94.4 & 93.7 & 93.1 & 92.2 & 92.2 \\ \hline
4 & 88.2 & 95.1 & 94.6 & 94.0 & 91.0 & 92.6\\ \hline
5 & 90.9 & 94.7 & 94.1 & 94.1 & 94.2 & 94.4\\ \hline
6 & 70.1& 83.0& 82.1 & 81.9 & 80.2 & 77.3\\ \hline \hline
\multicolumn{2}{|c|}{$\overline{\Delta \text{Accuracy}}$}&\textbf{+9.6}& \textbf{+8.2} & \textbf{+7.9} & \textbf{+6.8}& \textbf{+6.4}\\ \hline
\end{tabular}
\label{tab:found-rules}
\end{table}
}

In summary, MIRA presents a significant advancement over existing methods in terms of interpretability and customizability. It outperforms other \ac{rbs}s with its transparent and user-tailored approach, which is crucial for applications requiring a clear understanding of model decisions. Although a \ac{ffnn} might achieve better predictive performance, MIRA offers greater transparency, making it a preferable choice when explainability is a priority. The integration of personalized rules further accentuates the RBS's ability to fine-tune predictions to individual user characteristics, showcasing its adaptability and user-centric design.

\section{Conclusion}

This paper presents a significant advancement in the context of \ac{rbs}s, offering an interpretable solution tailored for radar-based gesture-sensing applications. We addressed the limitations of existing models and introduced novel methodologies for rule induction. By focusing on rule transparency and the incorporation of personalized rules, our \ac{rbs} MIRA not only meets the demands for interpretable AI but also provides a customized interface that adapts to individual users. The comparative evaluations and discussions based on an extensive \ac{fmcw} radar gesture dataset presented in this paper underscore the practicality and necessity of such systems.





\printbibliography


\end{document}